\documentclass[sigchi]{acmart}
\def\BibTeX{{\rm B\kern-.05em{\sc i\kern-.025em b}\kern-.08emT\kern-.1667em\lower.7ex\hbox{E}\kern-.125emX}}

\usepackage{float}
\usepackage{url}
\usepackage{multirow}
\usepackage{amsmath}
\usepackage{placeins}
\usepackage{array}
\usepackage{tablefootnote}
\usepackage{hyperref}

\usepackage{url}
\makeatletter
\g@addto@macro{\UrlBreaks}{\UrlOrds}
\makeatother

\newcolumntype{P}[1]{>{\centering\arraybackslash}p{#1}}

\renewcommand\footnotetextcopyrightpermission[1]{} 
\begin{document}

\title{Multi-label Classification for Automatic Tag Prediction in the Context of Programming Challenges}

\author{Bianca Iancu}
\affiliation{
  \institution{Delft University of Technology}
  \city{Delft}
  \state{South Holland}}
\email{A.Iancu-1@student.tudelft.nl}

\author{Gabriele Mazzola}
\affiliation{
  \institution{Delft University of Technology}
  \city{Delft}
  \state{South Holland}}
\email{G.Mazzola@student.tudelft.nl}

\author{Kyriakos Psarakis}
\affiliation{
  \institution{Delft University of Technology}
  \city{Delft}
  \state{South Holland}}
\email{K.Psarakis@student.tudelft.nl}

\author{Panagiotis Soilis}
\affiliation{
  \institution{Delft University of Technology}
  \city{Delft}
  \state{South Holland}}
\email{P.Soilis@student.tudelft.nl}

\begin{abstract}
One of the best ways for developers to test and improve their skills in a fun and challenging way are programming challenges, offered by a plethora of websites. For the inexperienced ones, some of the problems might appear too challenging, requiring some suggestions to implement a solution. On the other hand, tagging problems can be a tedious task for problem creators. In this paper, we focus on automating the task of tagging a programming challenge description using machine and deep learning methods. We observe that the deep learning methods implemented outperform well-known IR approaches such as tf-idf, thus providing a starting point for further research on the task. 
\end{abstract}

\maketitle

\section{Introduction}
\par In recent years, more and more people have started to show interest in competitive programming, either for purely educational purposes or for preparing for job interviews. Following this trend, we can also see an increase in the number of online platforms providing services such as programming challenges or programming tutorials. These problems are based on a limited number of strategies to be employed. Understanding which strategies to apply to which problem is the key to designing and implementing a solution. However, it is often difficult to directly infer from the textual problem statement the strategy to be used in the solution, especially with little or no programming experience. Thus, assisting the programmer by employing a system that can recommend possible strategies could prove very useful, efficient and educational. 
\par
\begin{figure}[h!]
    \centering
    \includegraphics[width=0.5\textwidth]{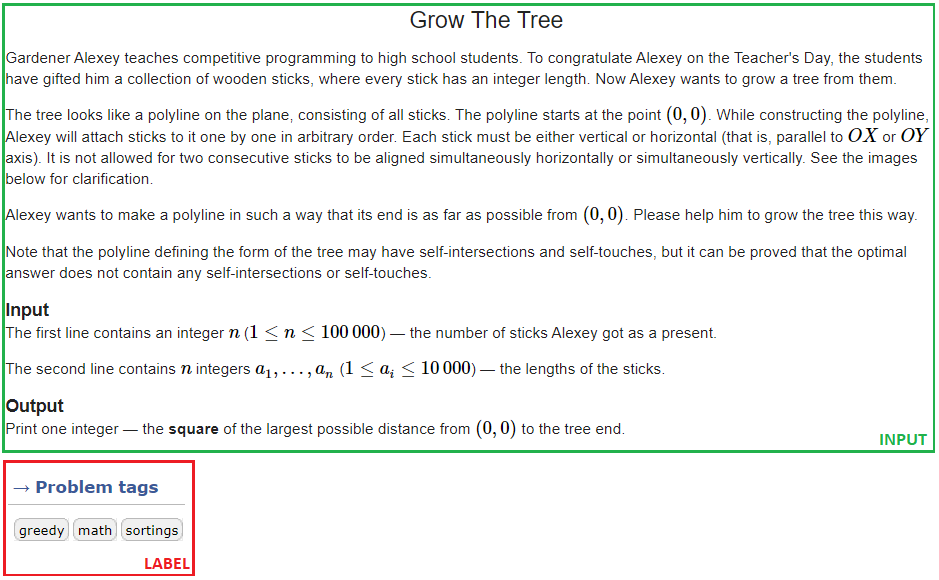}    \caption{Data sample example}
    \label{fig:training_sample}
\end{figure}

 To this end, we propose a system to automatically tag a programming problem given its textual description. These tags can be either generic, such as 'Math', or more specific, such as 'Dynamic Programming' or 'Brute Force'. Each problem can have multiple tags associated with it, thus this research is focusing on a multi-class multi-label classification problem. This is a more challenging problem than the usual multi-class classification, where each data point can only have one label attached. An example of a data sample for our problem is shown in Figure \ref{fig:training_sample}. A similar problem would be to predict these tags based on the source code of a solution, but this would restrict the application domain. To be more specific, we are interested in applying the system in the context of online programming challenges platforms where no solution or source code is available. Thus, we only consider the textual description of the problem statements as input to our system.

\par By gathering data from two of the main online programming challenges platforms, Topcoder \footnote{\url{https://www.topcoder.com/}} and CodeForces \footnote{\url{https://codeforces.com}}, we approach the problem through both machine learning and deep learning solutions, experimenting with different architectures and data representations. Considering the complexity of this problem, which is difficult even for humans, our hypothesis is that deep learning methods should be an effective way of approaching this problem, given enough data. Based on the aforementioned, the research question that we are trying to answer is the following: "\textit{Could deep learning models learn and understand what are the strategies to be employed in a programming challenge, given only the textual problem statement?}"

\par The rest of the paper is organized as follows: in Section \ref{sec:related_work} we describe the related work carried out in the literature, both regarding multi-label classification, as well as text representation methods. Subsequently, in Section \ref{sec:data} we describe the process of gathering, understanding and pre-processing the data, while in Section \ref{sec:methodology} we present the data representation methods and the models that we employ. Following that, in Section \ref{sec:experiments} we discuss the experimental setup and present the results, followed by a discussion and reflection regarding those in Section \ref{sec:discussion}. Finally, we conclude our research in Section \ref{sec:conclusion}.

\section{Related work} \label{sec:related_work}

\subsection{Multi-label classifcation}
\par As far as we are aware, no previous work has been carried out regarding multi-label classification in the context of tagging programming problem statements. Additionally, most of the papers tackling this task employ traditional machine learning methods, rather than deep learning approaches. In \cite{zhang2005k} the authors propose a multi-label lazy learning approach called ML-kNN, based on the traditional k-nearest neighbor algorithm. Moreover, in \cite{mccallum1999multi} the text classification problem is approached, which consists of assigning a text document into one or more topic categories. The paper employed a Bayesian approach into multiclass, multi-label document classification in which the multiple classes from a document were represented by a mixture model. Additionally, in \cite{katakis2008multilabel} the authors modeled the problem of automated tag suggestion as a multi-label text classification task in the context of the "Tag Recommendation in Social Bookmark Systems". The proposed method was built using Binary Relevance (BR) and a naive Bayes classifier as the base learner which were then evaluated using the F-measure. Furthermore, a new method based on the nearest-neighbor technique was proposed in \cite{huang2011multilabel}. More specifically, the multi-label categorical K-nearest neighbor approach was proposed for classifying risk factors reported in SEC form 10-K. This is an annually filed report published by US companies 90 days after the end of the fiscal year. 
\par A different strategy for approaching the multi-label classification problem proposed in the literature is active learning, examined in \cite{esuli2009active}. Furthermore, in \cite{yang2009effective} the authors proposed a multi-label active learning approach for text classification based on applying a Support Vector Machine, followed by a Logistic Regression.
\par Regarding deep learning approaches, in \cite{nam2014large} the authors analyzed the limitations of BP-MLL, a neural network (NN) architecture aiming at minimizing the pairwise ranking error. Additionally, they proposed replacing the ranking loss minimization with the cross-entropy error function and demonstrated that neural networks can efficiently be applied to the multi-label text classification setting. By using simple neural network models and employing techniques such as rectified linear units, dropout and AdaGrad, their work outperformed state-of-the-art approaches for this task. Furthermore, the research carried out in \cite{liu2017deep} analyzed the task of extreme multi-label text classification (XMTC). This refers to assigning the most relevant labels to documents, where the labels can be chosen from an extremely large label collection that could even reach a size of millions. The authors in \cite{liu2017deep} represented the first deep learning approach to XMTC using a Convolutional Neural Network. The authors showed that the proposed CNN successfully scaled to the largest datasets used in the experiments, while consistently producing the best or the second-best results on all the datasets.
\par When it comes to evaluation metrics in multi-label classification scenario, a widely employed metric in the literature is the Hamming loss \cite{katakis2008multilabel}\cite{read2008multi}\cite{trohidis2008multi}\cite{zhang2007ml}. Furthermore, more traditional metrics are also used such as the F-measure \cite{katakis2008multilabel}\cite{read2008multi}\cite{trohidis2008multi}, as well as the average precision \cite{trohidis2008multi}\cite{zhang2005k}.

\subsection{Text representation}
Apart from the aforementioned, a lot of literature work has gone into experimenting with different ways of representing text. According to \cite{turian_word_2010}, the word representation that has been traditionally used in the majority of supervised Natural Language Processing (NLP) applications is one-hot encoding. This term refers to the process of encoding each word into a binary vector representation where the dimensions of the vector represent all the unique words included in the corpus vocabulary. While this representation is simple and robust \cite{mikolov_efficient_2013}, it does not encode any notion of similarity between words. For instance, the word "airplane" is equally different to the word "university" as the word "student". The proposal of Word2Vec \cite{mikolov_efficient_2013} solves this issue by encoding words into a continuous lower dimensional vector space where words with similar semantics and syntax are grouped close to each other. This led to the proposal of more types of text embeddings such as Doc2Vec \cite{le_distributed_nodate} which encodes larger sequences of text, such as paragraphs, sentences, and documents, into a single vector rather than having multiple sectors, ie. one per word. One of the more recent language models which produced state-of-the-art results in several NLP tasks is the Bidirectional Encoder Representations from Transformers (BERT) \cite{devlin2018bert}. Its main innovation comes from applying bidirectional training of an attention model, the Transformer, to language modeling.
\section{Data} \label{sec:data}
In this section, we describe the steps carried out to obtain the dataset that we have worked with. More specifically, we focus on presenting the data source and the data collection process, followed by an overview of the preprocessing that we perform. Additionally, we show several descriptive statistics regarding the data and explain the steps we take for defining the tag taxonomy for the dataset.


\subsection{Data sources \& collection}
We investigate different competitive programming platforms to build a dataset of programming statements and tags. Specifically, the platforms of interest to us are: TopCoder\footnote{\url{https://www.topcoder.com/}}, Hackerrank \footnote{\url{https://www.hackerrank.com/}}, CS Academy\footnote{\url{https://csacademy.com/}}, Codewars\footnote{\url{https://www.codewars.com/}}, and Codeforces\footnote{\url{http://codeforces.com/problemset}}. Due to legal reasons as well as the complexity of the platform interfaces, we manage to successfully scrape only two of these platforms:
\begin{itemize}
    \item Codeforces on 13/09/2019. Codeforces is a website that offers programming contests to people interested in participating. Users are free to upload a challenge statement, together with several tags specifying the strategies that should be used to devise a solution. We scrape a total of 5,341 problems, together with their tags.
    \item Topcoder on 17/09/2019. Topcoder is a crowdsourcing platform in the sense that each problem is made available to all the developers and they provide solutions. It also provides an archive with free access, containing a set of problems with tags, similarly to CodeForces. We scrape a total of 4,508 problems, together with their tags.
\end{itemize}
We implement the scraping of these two platforms in Python, creating two ad-hoc scripts, one per platform. We use BeautifulSoup4\footnote{\url{https://pypi.org/project/beautifulsoup4/}} to parse the HTML of the pages, to extract only meaningful information.

\subsection{Data preprocessing}
The crawled data contain a lot of redundant information or even data that can impact the training process in a negative way (e.g. HTML tags and \LaTeX symbols). What is more, some NLP techniques, such as stop-word removal, have proven to increase classification performance on textual data since they do not provide any additional information and only increase the dimensionality of the problem \cite{silva2003importance}. Following the collection process, we merge the two crawled datasets and preprocess them.

At first, HTML tags $(<.*?>)$ are removed since they do not provide any descriptive information about the problem. Afterward, since mathematical definitions are in \LaTeX format on CodeForces and as plain text on TopCoder we decide to remove them from both to avoid introducing differences between the two sets of data. In addition, all non-ASCII characters, digits, stop-words, punctuation and one character words (i.e. variable names) are removed for the same reason.

The next step of the preprocessing pipeline is to concatenate all textual fields into one (i.e. title and description are concatenated into the field "text"). Furthermore, we convert every character in this new field to lowercase. Finally, we run an exact duplicate removal since we observed that some challenges on both crawled websites appear more than once. Then we remove words with less than 10 occurrences in the entire corpus assuming that they are not descriptive enough. Thus, the final dataset is a JSON file with an array of coding challenges and each entry has two fields, the "text" and the "tags" associated with it. 

\subsection{Descriptive statistics}
\par After performing the preprocessing steps, we compute several descriptive statistics. More specifically, we compute the total number of problems, the average word count per problem and the average tags per problem, all separately for the CodeForces and Topcoder data, as well as for the combined dataset. These statistics are provided in Table \ref{tab:desc_stats}.

\begin{table}[h!]
\centering
\resizebox{\columnwidth}{!}{%
\begin{tabular}{cccc}
\hline
\textbf{} & \textbf{CodeForces} & \textbf{Topcoder} & \textbf{Combined dataset} \\ \hline
\textit{Problem count} & 4,592 & 4,115 & 8,707 \\ \hline
\textit{Avg word count/problem} & 120.93 & 94.81 & 108.58 \\ \hline
\textit{Avg tags/problem} & 1.76 & 1.46 & 1.62 \\ \hline
\end{tabular}%
}
\caption{Descriptive statistics}
\label{tab:desc_stats}
\end{table}

\par As we can observe, the number of problems gathered from the two data sources, as well as the additional statistics, are quite balanced. There are no significant differences when it comes to the average word count per problem and the average tags per problem between the CodeForces data and the Topcoder data. For a breakdown of the average number of words per tag, the reader can refer to Appendix A.

\subsection{Tag taxonomy}

Since each of the websites uses a different set of tags, it is necessary to create a taxonomy that maps the original tags to new common target labels. To decide on the class labels to be included in the taxonomy, we start by analyzing the data. To this end, we generate a correlation matrix for all the original tags present in the dataset to infer which of them can be grouped in a more general label. The original number of labels was 16 for the Topcoder data and 35 for the CodeForces data. After performing the label aggregation we result in 17 common tags for the two data sources. For a complete list of these, the reader can refer to Appendix B.
\par
\begin{figure}[h!]
    \centering
    \includegraphics[width=0.5\textwidth]{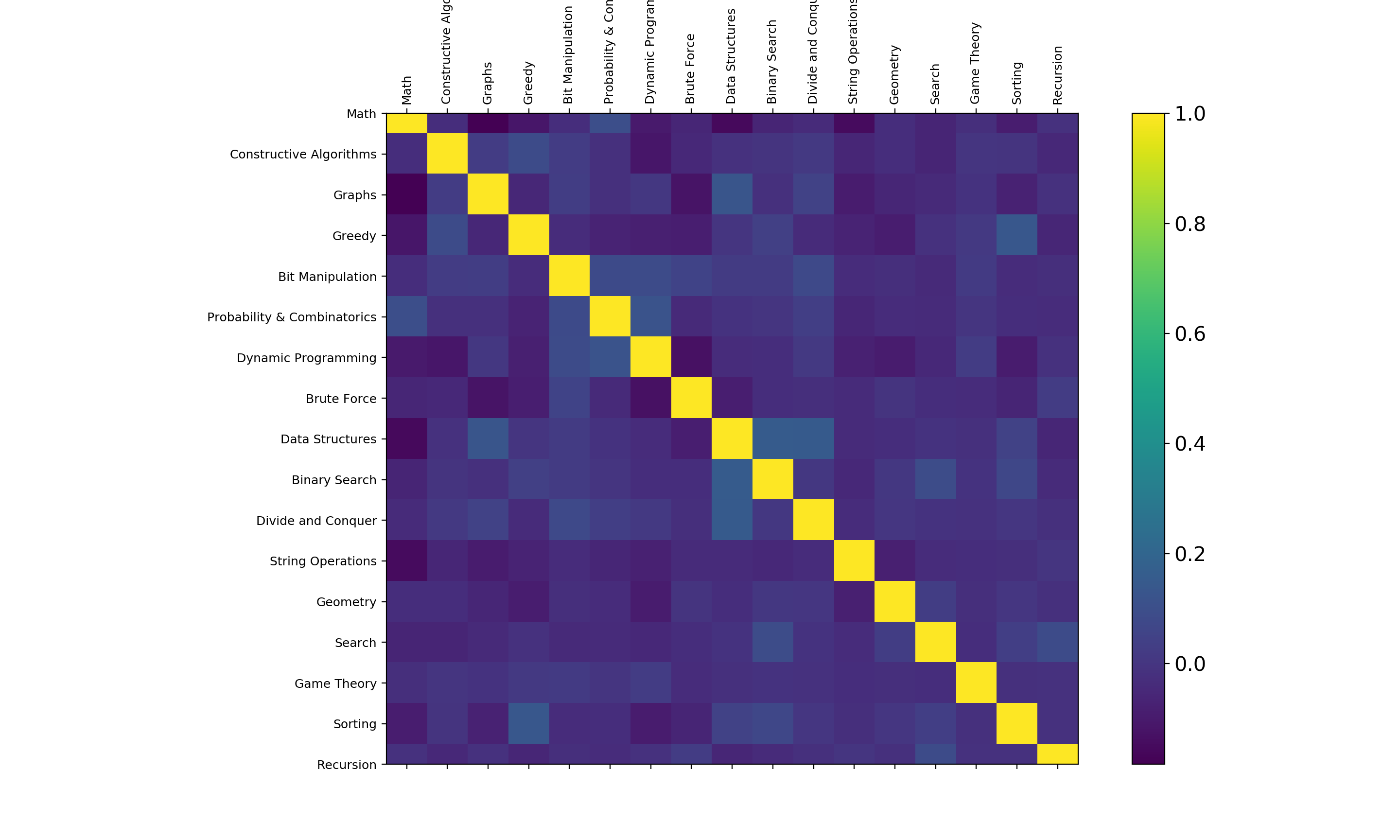}
    \caption{Problem tag correlation}
    \label{fig:corr_17}
\end{figure}

 Following preliminary experimentation, we observe that the performance is unsatisfactory which brings us to suspect that we have too few data instances given the associated number of labels. Thus, we proceed by further aggregating and reducing the number of tags. As before, we perform a visual analysis of the correlation matrix in Figure \ref{fig:corr_17} between the different tags to decide which of them can be aggregated. 

\begin{figure}[h!]
    \centering
    \includegraphics[width=0.5\textwidth]{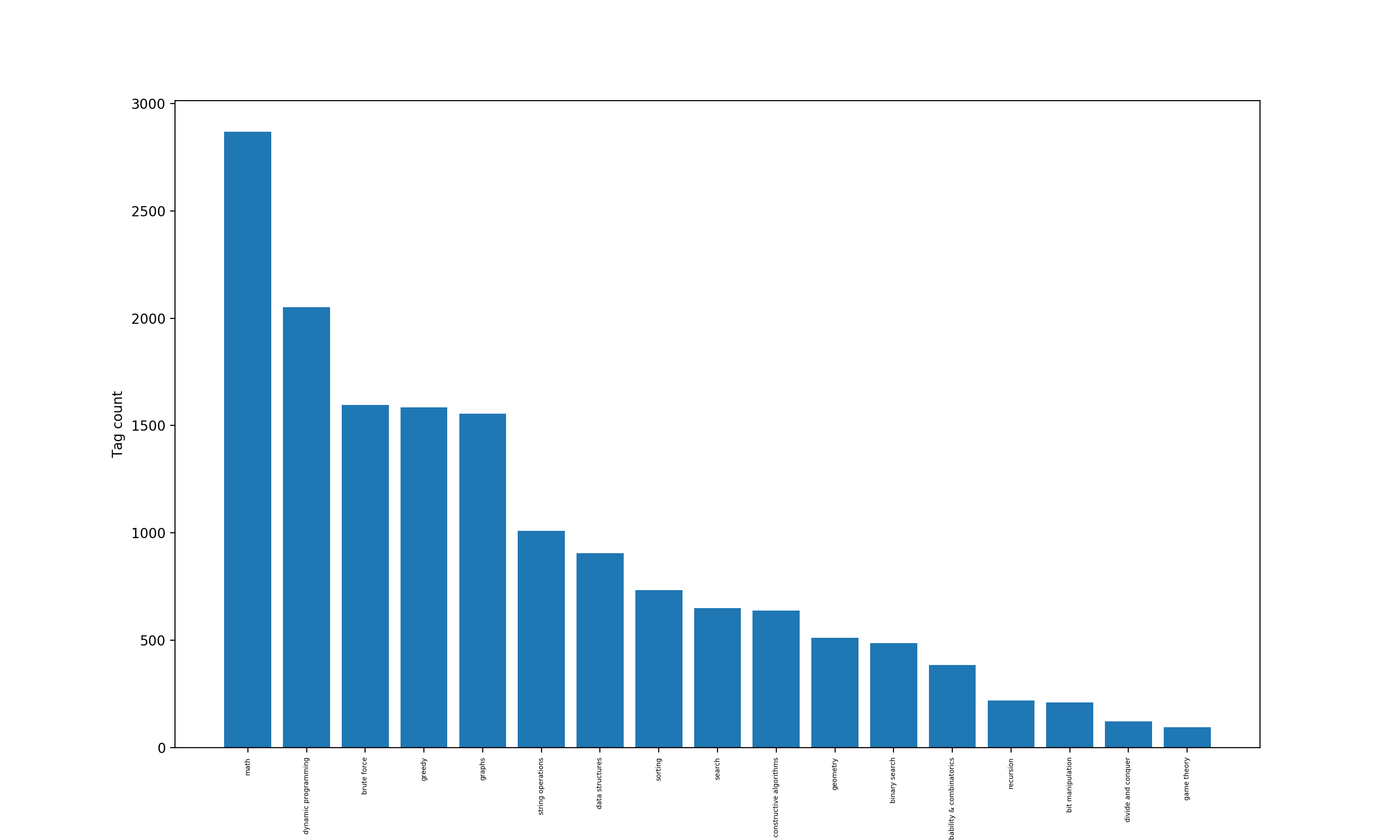}
    \caption{Distribution of the original taxonomy}
    \label{fig:taxonomy_orig_distr}
\end{figure}

Additionally, we plot a bar chart of the frequencies of the class labels in Figure \ref{fig:taxonomy_orig_distr} to understand whether some classes are present in too few instances to be properly learned by the models.

After performing this additional aggregation, we decrease the number of tags from 17 to 9.  For a complete list of these, the reader can refer to Appendix C. The main semantic decisions taken while creating the final taxonomy for the data are the following:
\begin{enumerate}
    \item We remove several general tags since we consider that they are too general and vague and, thus, do not give any concrete information regarding the methodology associated with the problem statement. Such tags are, for example, 'implementation', 'programming', 'arrays' or 'interactive'.
    \item We group problem statements originally labeled with tags related to strings, string manipulations, and regular expressions into one category under the 'String Operations' name.
    \item The term 'ternary search' was found to correspond to similar operations as 'binary search'\footnote{\url{https://www.geeksforgeeks.org/ternary-search/}}. Thus, we group their associated problem statements into one label, 'Binary Search'. During the second aggregation stage, we further combine 'Binary Search' with the 'Search' label under the common name 'Search and Binary Search', as although they use different methods, they refer to a similar class of problems. 
    \item Since we observed that tags like 'shortest path' and 'dfs' were associated with both 'graphs' and 'trees', we merge the latter two into the 'Graphs' label. Additionally, since graphs are a data structure, we further combine them with 'Data Structures' in the second aggregation stage into the 'Data Structures and Graphs' label.
    \item Regarding the 'Math' tag we initially intended to remove it, since we consider it to be quite general. However, after further analysis we observe that it is present in a high number of data points, hence we decide to keep it in the taxonomy. In the second aggregation stage, we combine it with the 'Probabilities and Combinatorics' tags, since they are a mathematics field as well, under the common name 'Math and Probabilities'.
\end{enumerate}

\par
\begin{figure}[h!]
    \centering
    \includegraphics[width=0.5\textwidth]{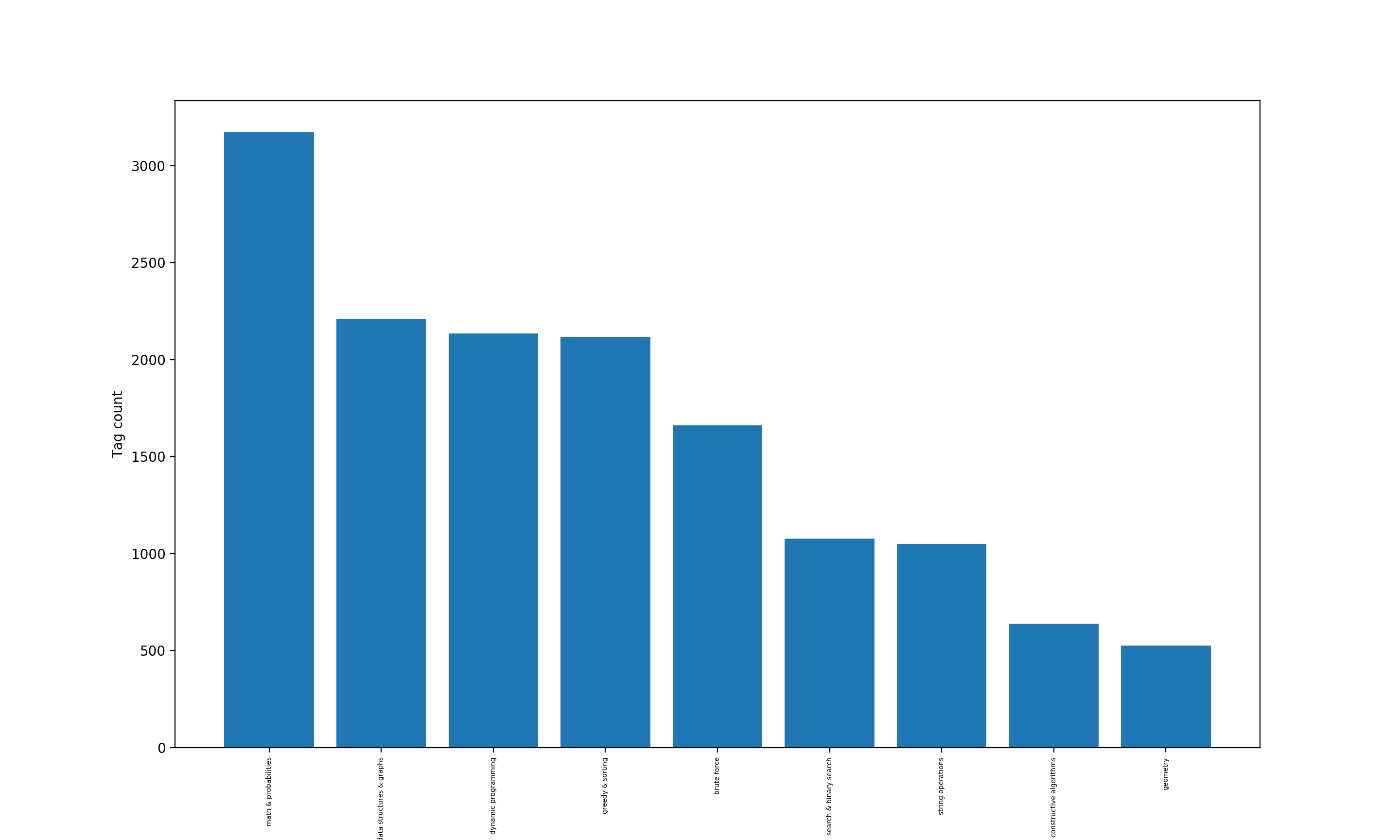}
    \caption{Distribution of the final taxonomy}
    \label{fig:taxonomy_final_distr}
\end{figure}

 The distribution of the final taxonomy associated with our dataset is shown in Figure \ref{fig:taxonomy_final_distr}. We can observe that there is an imbalance present in the dataset, with approximately 2500 data points difference between 'Math and Probabilities' (the most common tag) and 'Geometry' (the least common tag). The imbalance is further taken into account when implementing the machine learning and deep learning models. Additionally, it is important to note that while we define 9 classes for classifying the data points, according to the descriptive statistics presented before, in our data there are less than 2 tags per data point on average. Thus, the models will have to learn to predict significantly less 1s than 0s for the associated labels. Therefore, to account for this matter, in the methodology presented in the next section, we weigh the evaluation metric, as well as the loss function for the deep learning models to give more importance to correctly predicting 1s during the training.

\section{Methodology} \label{sec:methodology}
In this section, we first describe the data representation techniques that we used to encode our data. Following that, the machine and deep learning techniques used to classify the programming problem statements into the target set of tags are outlined. It has to be underlined that we only provide the intuition and main concepts of these methods since they do not constitute the main focus of our work.

\subsection{Data representation}

\subsubsection{tf-idf}   
To obtain features for our baseline classifier we employ the \textbf{tf-idf} approach. The Python library that we used\footnote{\url{https://scikit-learn.org/stable/modules/feature\_extraction.html}} employs the following formula: \[\text{tf-idf(t,d)} = \text{tf(t,d)} * \text{idf(t)}\] where $\text{tf(t,d)}$ is the number of times term $t$ appears in document $d$, and $\text{idf(t)} = log\frac{1+n}{1+df(t)} + 1$ where $n$ is the total number of documents and $\text{df(t)}$ is the number of documents containing term $t$. By doing this, we obtain a number of features for each document equal to the total number of unique terms. For a more detailed overview of tf-idf and its usage we refer the reader to \cite{ramos2003using}.

\subsubsection{One-hot encoding}
Another representation technique we experiment with is one-hot encoding. This method represents textual data in discrete vectors with binary values. As mentioned previously, one-hot encoding treats all the words in equal fashion and does not preserve any information about the word meaning. Another thing to note is that it also leads to a feature space with very high dimensionality. In particular, the number of dimensions of the one-hot vector is equal to the number of unique words in the programming problem statements dataset.

\subsubsection{Word2Vec}
One of the most popular representation techniques in the field of NLP is the Word2Vec \cite{mikolov_efficient_2013} model. In particular, in their work, the authors propose two different feedforward networks, Continuous Bag-of-Words (CBOW) and Skip-gram, that project words into a continuous vector space which preserves their semantics and syntax. The main difference between the two models is that CBOW attempts to predict the target word based on its context, whereas Skip-gram predicts the context of the current word. In our work, we used the Skip-gram method since it was found to outperform CBOW in \cite{mikolov_efficient_2013}. More details can be found in the original Word2Vec paper \cite{mikolov_efficient_2013}.

\subsubsection{Doc2Vec}
The final representation that was tried out is the Doc2Vec embeddings proposed by \cite{le_distributed_nodate}. In simple terms, it extends the continuous vector space idea from Word2Vec \cite{mikolov_efficient_2013} to larger text sequences, such as sentences, paragraphs, and documents. To be more exact, we map each programming problem statement to a pre-defined number of dimensions. Similarly to Word2Vec, two models are proposed called Distributed Memory (PV-DM) and Distributed Bag of Words (PV-DBOW) which correspond to the CBOW and Skip-gram from Word2Vec \cite{mikolov_efficient_2013} respectively. In this paper, we use the PV-DBOW model to obtain the document embeddings. More information about Doc2Vec are available in \cite{le_distributed_nodate}.

\subsection{Models}
\par For carrying out the multi-label classification, we employ both machine learning and deep learning approaches, thus comparing the performances and analyzing the influence of deep learning on this task.
\par The \textbf{machine learning} models we use for classification are the following:
\begin{enumerate}
    \item \textit{Decision Tree}: a non-parametric supervised learning technique that performs classification by learning decision rules based on the data features. The model is implemented using the \textit{scikit-learn}\footnote{\url{https://scikit-learn.org/stable/modules/tree.html}} Python library.
    \item \textit{Random Forest}: a model that fits several decision tree classifiers on different sub-samples of the data and averages their outputs. It is also implemented using the \textit{scikit-learn}\footnote{\url{https://scikit-learn.org/stable/modules/generated/sklearn.ensemble.RandomForestClassifier.html}} Python library.
    \item \textit{Random Classifier}: We implement our multiclass multilabel random classifier by predicting vectors where each entry has an equal probability of being either zero or one. We, therefore, expect to have, on average, 4.5 ones predicted per sample.
\end{enumerate}

\par The \textbf{deep learning} model that we employ is a \textit{Long-Short Term Memory (LSTM)} network, which is an improvement to the traditional Recurrent Neural Networks (RNNs). By using the Gating mechanism, it addresses the short-term memory challenge that RNNs have. Thus, LSTMs can preserve long-term memory, which is an important aspect to consider since we are dealing with a text classification task. When generating the labels, it is essential to preserve the long-term dependencies in the problem statements. In order to obtain the final predicted labels we use a logistic activation function. The reason that logistic activation function was used over softmax is that the latter is more commonly used for multi-class classification problems where there is only one target label. We implement the LSTM architectures using PyTorch\footnote{\label{PyTorch}\url{https://pytorch.org/}}, a Python-based open-source deep learning library.
\section{Experiments} \label{sec:experiments}
\par In this section, we provide an overview of the experimental setup. More specifically, we discuss the training and test sets, the evaluation metrics that we employ, as well as the training setup and the model hyperparameters. Additionally, we present the experimental results.

\subsection{Training \& Test sets}
Being aware of the fact that we have little data available, 8,707 data points after pre-processing and duplicate removal, we decide to perform a single split between training and test, without creating a validation set. We perform a 90-10 split, creating a training set consisting of \textbf{7826} training samples, and \textbf{881} test samples. This decision is backed up by the fact that (i) it is not straightforward to balance the sets since we are in a multilabel setting (one problem can have zero, one, or multiple tags) and (ii) our goal is not devising a state-of-the-art architecture for the given problem, but rather verifying the applicability of neural networks to tackle it.
\par We mitigate the effect of this decision by avoiding extensive hyperparameter tuning using the test set, which would otherwise cause our model to overfit on the test set\footnote{\url{https://www.kdnuggets.com/2019/05/careful-looking-model-results-cause-information-leakage.html}}.
Therefore, the results we present later on in this section are obtained on the test set: the models do not have access to this data, but we use it to make decisions and tune a limited number of hyperparameters.

\subsection{Evaluation Metrics}
We present here the evaluation metrics we adopt to evaluate the different models.
\subsubsection{Weighted Hamming Score}
The standard (non-weighted) Hamming Score is defined as "$1 - HammingLoss$", where the Hamming Loss represents the fraction of the wrong labels out of the total number of labels. However, as mentioned before, the datapoints have, on average, less than 2 tags (out of 9). For this reason, we decide to weight the Hamming Score metric by weighting differently the errors in predicting 1s or 0s. The implementation of our custom metric is as follows:
\[\scriptstyle Weighted Hamming Score = 1 - Weighted Hamming Loss\] where \[\scriptstyle Weighted Hamming Loss = W_1 * Ratio\_Miscl_1 + W_0 * Ratio\_Miscl_0\]
and $Ratio\_Miscl_i$ is the ratio of misclassified entries per label \textit{i}. $W_0$ and $W_1$ are set to 0.18 and 0.82, respectively, to account for the label unbalance present in our dataset.

\subsubsection{Average Precision, Recall, F1}
\par We implement the average precision, recall, and F1 score using the sklearn.metrics library \footnote{\url{https://scikit-learn.org/stable/modules/generated/sklearn.metrics.average\_precision\_score.html}} \footnote{\url{https://scikit-learn.org/stable/modules/generated/sklearn.metrics.recall\_score.html}} \footnote{\url{https://scikit-learn.org/stable/modules/generated/sklearn.metrics.f1\_score.html}}. Our implementation weights the average obtained per label based on its support.

\begin{table*}[h!]
\centering
\resizebox{\textwidth}{!}{
\begin{tabular}{|c|c|c|c|c|c|c|c|}
\hline
\textbf{Model}     & \textbf{Weighted Hamming Score} & \textbf{Avg Precision} & \textbf{Avg Recall} & \textbf{Avg F1} & \textbf{Loss}  & \textbf{Avg \# of Ones per sample} & \textbf{N. of trainable params}\\ \hline
tf-idf + Random Forest & 0.27                           & 0.719                  & 0.111               & 0.171           & -              & 0.22 & -                              \\
tf-idf + Dec. tree & 0.423                           & 0.326                  & 0.331               & 0.323           & -              & 1.68 & -                              \\
Random classifier  & 0.518                           & 0.228                  & 0.525               & 0.309           & -              & 4.62  & -                             \\
one-hot + LSTM       & 0.708                           & \textbf{0.407}         & 0.7                 & \textbf{0.502}  & 0.163          & 2.61   & 402,009                            \\
Doc2Vec + FFNN       & 0.754                           & 0.387                  & 0.79                & 0.483           & 0.173          & 3.75 &  649                              \\
Word2Vec + LSTM         & \textbf{0.762}                  & 0.375                  & \textbf{0.793}      & 0.489           & \textbf{0.158} & 4 & 20,505 \\ 
\hline
\end{tabular}
}
\caption{Model results on test set}
\label{tab:classifier_results}
\end{table*}

\subsubsection{Average Number of Ones per Sample} 
As an additional metric, we keep track of the average number of ones predicted per sample by the model. We do this mainly to make sure our models are not predicting either all ones or all zeros, and to account for the label unbalance of the dataset.

\subsection{Training}
As explained earlier in this section, we perform a 90-10 split of our data, obtaining training and test set. The test set is only used to perform early stopping (to halt the training if the model does not improve on the test set for more than 10 epochs) and to tune a very limited number of hyperparameters. To obtain a test set with distribution as similar as possible to the training set, we employ the iterative stratification approach\footnote{\url{http://scikit.ml/api/skmultilearn.model\_selection.iterative\_stratification.html}}, which aims to solve the problem of balancing the split in the case of a multiclass multilabel setting.
\par We implement our models using the deep learning framework Pytorch\textsuperscript{\ref{PyTorch}} and using some of the already implemented models of scikit\footnote{\url{https://scikit-learn.org/stable/supervised\_learning.html}}. All of our code can be found on GitHub\footnote{\url{https://github.com/serg-ml4se-2019/group11-tagging-algorithm-problems}}. 
\par For the training of our architectures we employ binary cross-entropy loss \footnote{\url{https://pytorch.org/docs/stable/nn.html\#bceloss}}. However, similarly to the customization performed on the Hamming Loss, we weight the binary cross-entropy loss to account for the label unbalance.

\subsection{Model hyperparameters}
With regards to the\textbf{ tf-idf} representation, the input dimensionality is \textbf{6,259}, which is equal to the number of unique terms in the entire corpus. For both \textbf{Decision Tree} and \textbf{Random Forest} we employ the default values of Sklearn, and set the number of estimators for the Random Forest classifier to \textbf{500}.
\par Similarly, with one-hot encoding, the input dimensionality equals the number of uniques terms in the corpus (\textbf{6,259}). The best performing LSTM network for this data representation has \textbf{16} hidden units, followed by a fully connected layer for the output classification (\textbf{9} classes). We use zero-padding on the input sequences to create batches and speed up training. The learning rate is \textbf{0.01}.

\par The \textbf{Doc2Vec} approach results in a smaller input representation, with a dimensionality of \textbf{30}. The architectures has two fully connected layers, with \textbf{16} and \textbf{9} neurons respectively. The non-linear activation function used is \textbf{ReLu}. The learning rate is again \textbf{0.01}.

\par The \textbf{Word2Vec} data representation results in an input representation of \textbf{300}. The LSTM has \textbf{16} hidden units and it is followed by a fully connected layer that maps the hidden state to an output of dimensionality \textbf{9}, for the final classification. Again, we employ zero-padding on the input sequences. The learning rate is \textbf{0.005}.

\subsection{Results}
Table \ref{tab:classifier_results} shows the performance of our models on the test set. We want to highlight again that although this data has been used for early stopping and limited tuning, the models have never seen it during training. It can be seen that \textbf{W2V + LSTM} achieves the best Weighted Hamming Score which is, in general, higher for the neural network models compared to the two baselines found in the first three rows of the table. 
\section{Discussion \& Reflection} \label{sec:discussion}
In this section we discuss our study, the findings arising from our experiments and the challenges faced while conducting them. 
\par To our knowledge, this is the first research work that focuses on predicting tags based on the textual description of programming problem statements. As a result, we cannot compare the performance we obtain with any state-of-the-art result. 
\par When it comes to interpreting the results shown in Table \ref{tab:classifier_results}, we consider as baselines the tf-idf with Decision Tree, since it achieves better performance than Random Forest in terms of the Weighted Hamming Score and the Random Classifier. We can observe that the deep learning models achieve significantly higher performance compared to our baselines, performing better on all the considered metrics. Regarding the data representation methods employed for the deep learning models, Word2Vec outperforms both Doc2Vec and one-hot encoding. This is a sensible result, as Word2Vec preserves the semantics and syntax of the words, as opposed to one-hot encoding. Additionally, it allows for the sequential modeling of the words in the problem statements by employing an LSTM, as opposed to Doc2Vec which encodes the data at a paragraph level. However, as we can observe in Table \ref{tab:classifier_results}, the number of trainable parameters is considerably higher for the Word2Vec model than for the Doc2Vec one, while only achieving a slight improvement in performance. Thus, depending on the use case, Doc2Vec might be preferred over Word2Vec. For instance, if a system needs to be retrained regularly to account for new data, using Doc2Vec instead might be a reasonable choice. Nevertheless, both the Word2Vec and Doc2Vec models have a substantially lower number of trainable parameters compared to one-hot encoding.
\par Moreover, we notice that in the case of the W2V+LSTM model, our best performing one in terms of Weighted Hamming Score, the average predicted number of ones per sample is quite high in comparison to the true labels (4 compared to 1.62). This could potentially be solved by adding a custom regularization term to this model to force the number of predicted ones to remain lower. 
\par Additionally, during training, we have observed that our deep learning models were overfitting on the training set after a few epochs, even after experimenting with regularization techniques such as drop-out and weight decay. Our reasoning behind this issue is that we do not have enough data since the problem that we aimed at solving serves a very specific purpose. This observation is also apparent in our baseline creation where more complex classifiers like Random Forest perform worse than a simpler Decision Tree. For an example of the overfitting problem, we refer the reader to Appendix D, where we include a training curve showing an overfitting behaviour (one-hot encoding + LSTM). That said, we notice that there is an emerging trend in deep learning literature that aims at solving problems that cannot be represented by large amounts of data, called small sample learning \cite{shu2018small}. In particular, models try to learn concepts rather than patterns since a concept can generalize better, thus avoiding overfitting. Such an approach could be investigated as future work for this task.
\section{Conclusion} \label{sec:conclusion}
In this work we investigated how deep learning techniques perform in a novel multi-class multi-label text classification problem. Our findings show that deep learning approaches significantly outperform traditional widely accepted IR techniques like tf-idf. Moreover, we experimented with different combinations of text representations and neural network architectures, finding Word2Vec + LSTM as the option that yields the best performance. Finally, we were able to experience first hand the challenges and issues arising when having a limited amount of data, an issue that is common in the deep learning literature.

Regarding possible future research directions, the need to collect more data for this task is of crucial importance. This could be done by either gathering data from more programming challenges websites or slightly different domains that can provide similar data samples. Another way to get more training data is by artificially augmenting the dataset either by the use of synonyms or adversarial networks \cite{gupta2019data}\cite{yu2017seqgan}. The availability of more training data will then allow researchers to use more advanced text embeddings such as BERT \cite{devlin2018bert} and XLNet \cite{yang2019xlnet} which are the current state-of-the-art in the NLP field. Last but not least, different lines of work can also be tried such as avoiding preprocessing the dataset and learning character embeddings, similar to what was performed in \cite{gelman2018language}.

\bibliographystyle{plain}
\bibliography{main}

\newpage
\section*{Appendix}

\section*{A. Dataset Descriptive Statistics}

\begin{figure}[h!]
    \centering
    \includegraphics[width=0.45\textwidth]{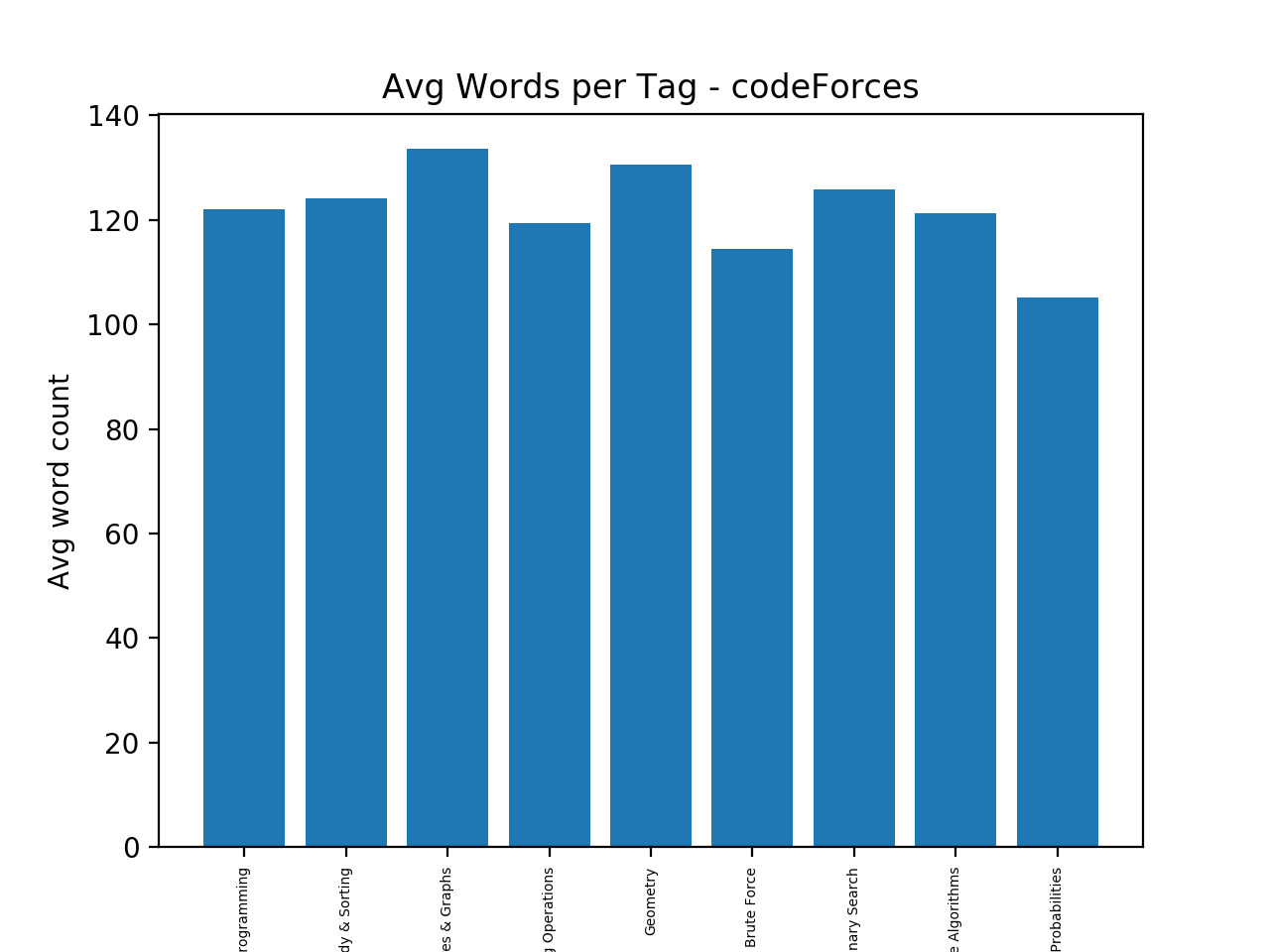}
    \caption{Average number of words per tag for the CodeForces data}
    \label{fig:avg_words_per_tag_codeforces}
\end{figure}

\begin{figure}[h!]
    \centering
    \includegraphics[width=0.45\textwidth]{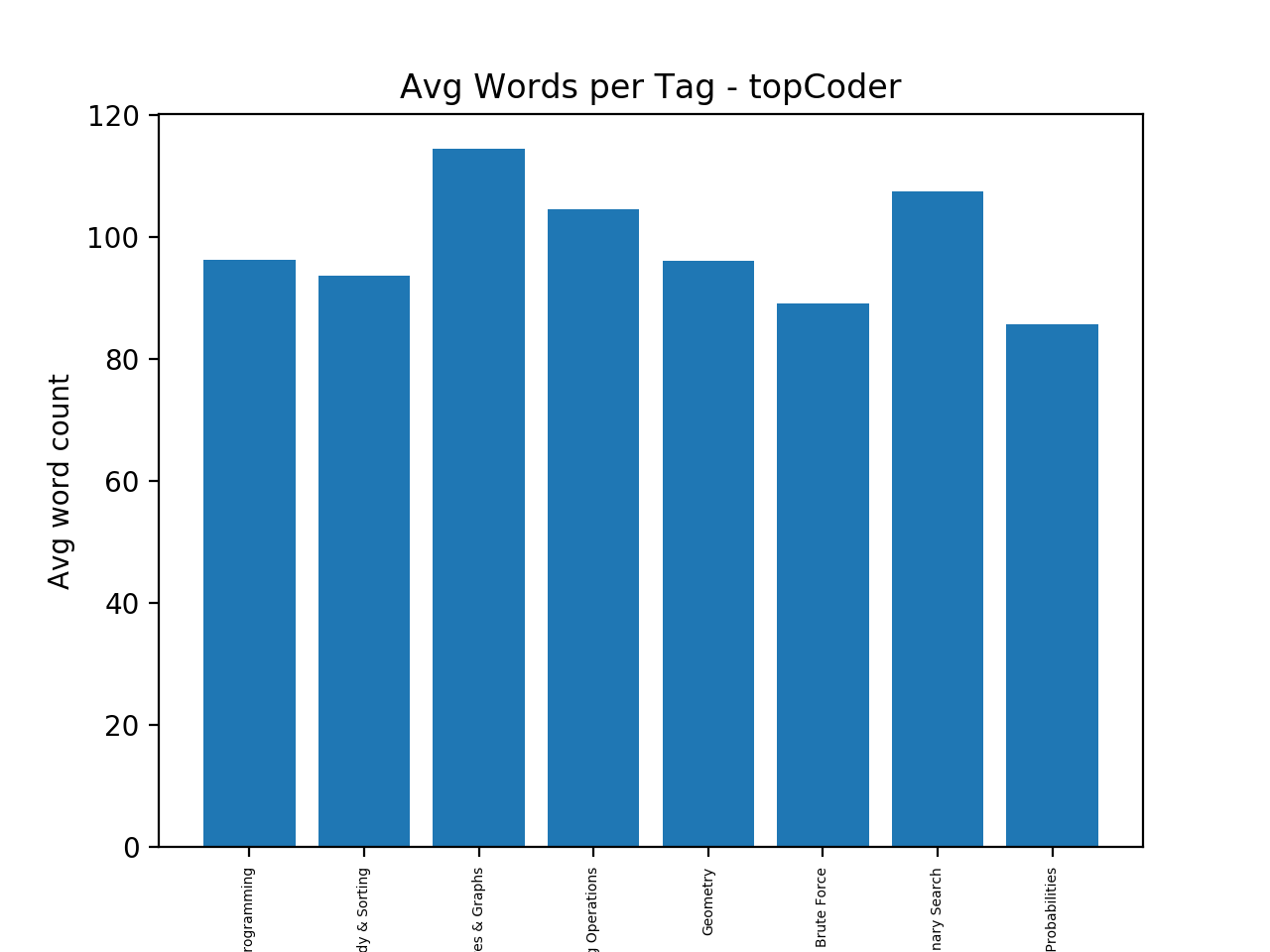}
    \caption{Average number of words per tag for the Topcoder data}
    \label{fig:avg_words_per_tag_topcoder}
\end{figure}

\begin{figure}[h!]
    \centering
    \includegraphics[width=0.45\textwidth]{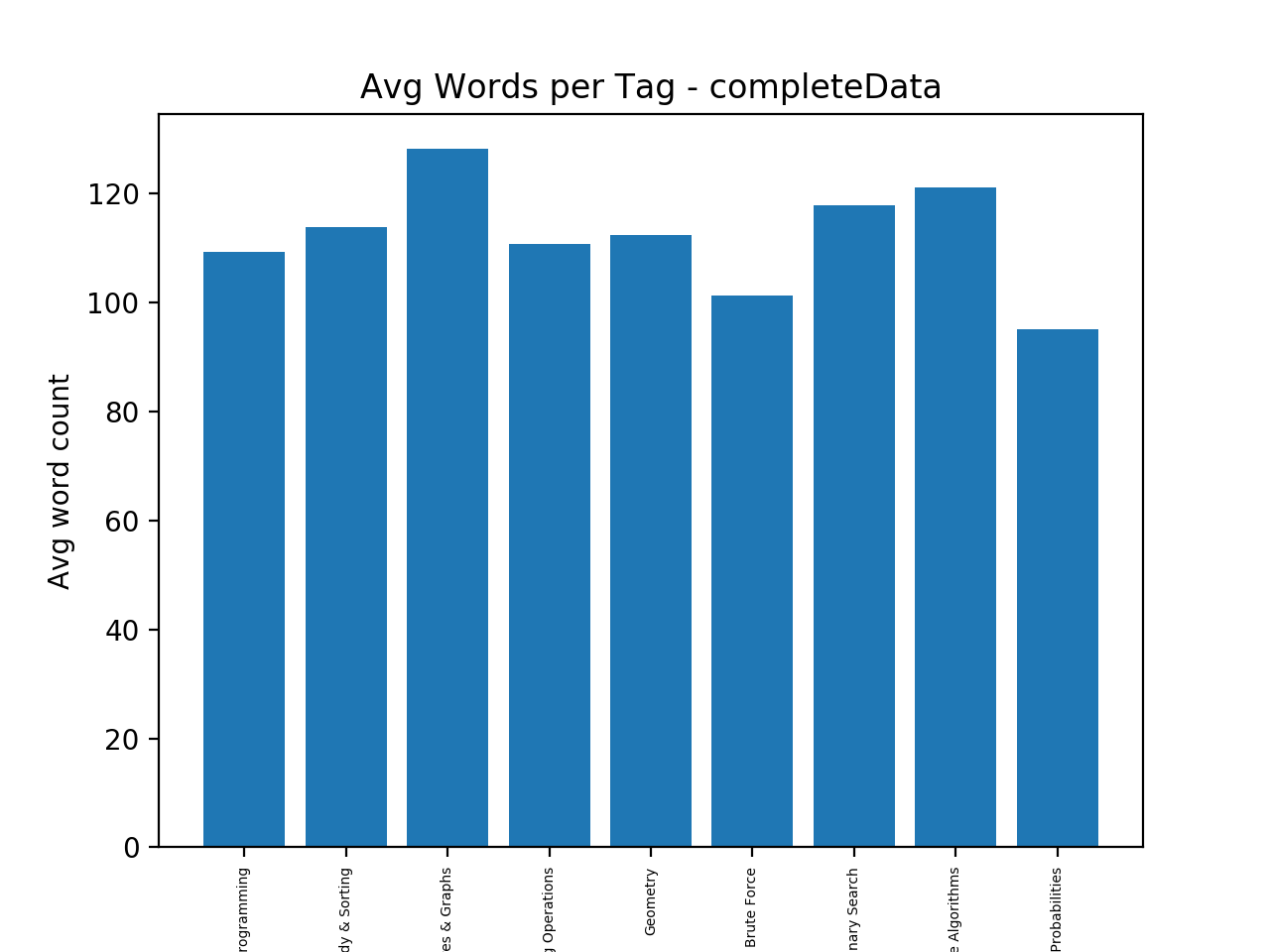}
    \caption{Average number of words per tag for the full combined data}
    \label{fig:avg_words_per_tag_full_data}
\end{figure}

\section*{B. List of initial 17 tags} \label{appendix:original_tags}
\par After the initial aggregation of the tags present in the two datasets, scraped from Topcoder and CodeForces, the 17 resulting problem statements tags are the following:
\begin{enumerate}
    \item Dynamic Programming
    \item Greedy
    \item Sorting
    \item Recursion
    \item Graphs
    \item String Operations
    \item Data Structures
    \item Divide and Conquer
    \item Geometry
    \item Bit Manipulation
    \item Brute Force
    \item Binary Search
    \item Search
    \item Game Theory
    \item Constructive Algorithms
    \item Math
    \item Probabilities and Combinatorics
\end{enumerate}

\section*{C. List of final 9 tags} \label{appendix:final_tags}
\par After the final aggregation of the initial 17 tags, the 9 final problem statement tags are the following:

\begin{figure}[h!]
    \centering
    \includegraphics[width=0.45\textwidth]{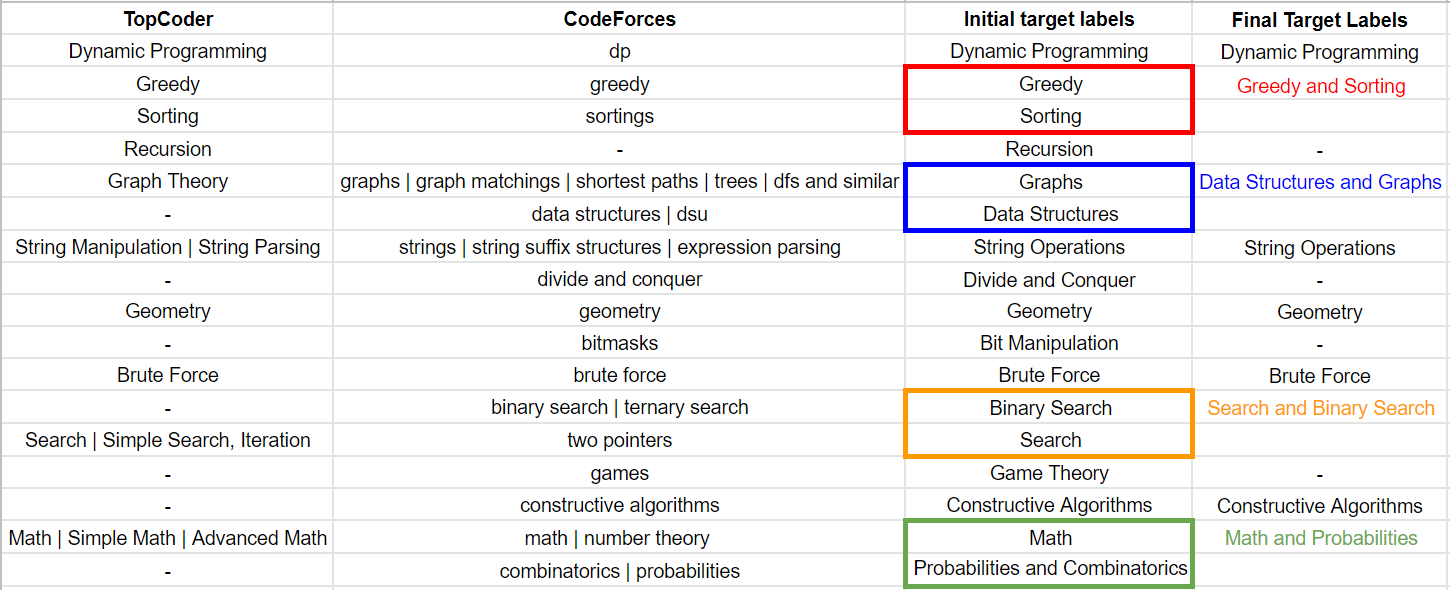}
    \caption{Final taxonomy}
    \label{fig:final_taxonomy}
\end{figure}

\begin{enumerate}
    \item Dynamic Programming
    \item Greedy and Sorting
    \item Data Structures and Graphs
    \item String Operations
    \item Geometry
    \item Brute Force
    \item Search and Binary Search
    \item Constructive Algorithms
    \item Math and Probabilities
\end{enumerate}

\par Figure \ref{fig:final_taxonomy} shows the aggregation that was performed on the initial 17 common tags in order to reduce them to the final 9 tags.

\section*{D. Overfitting problem}
\begin{figure}[h!]
    \centering
    \includegraphics[width=0.45\textwidth]{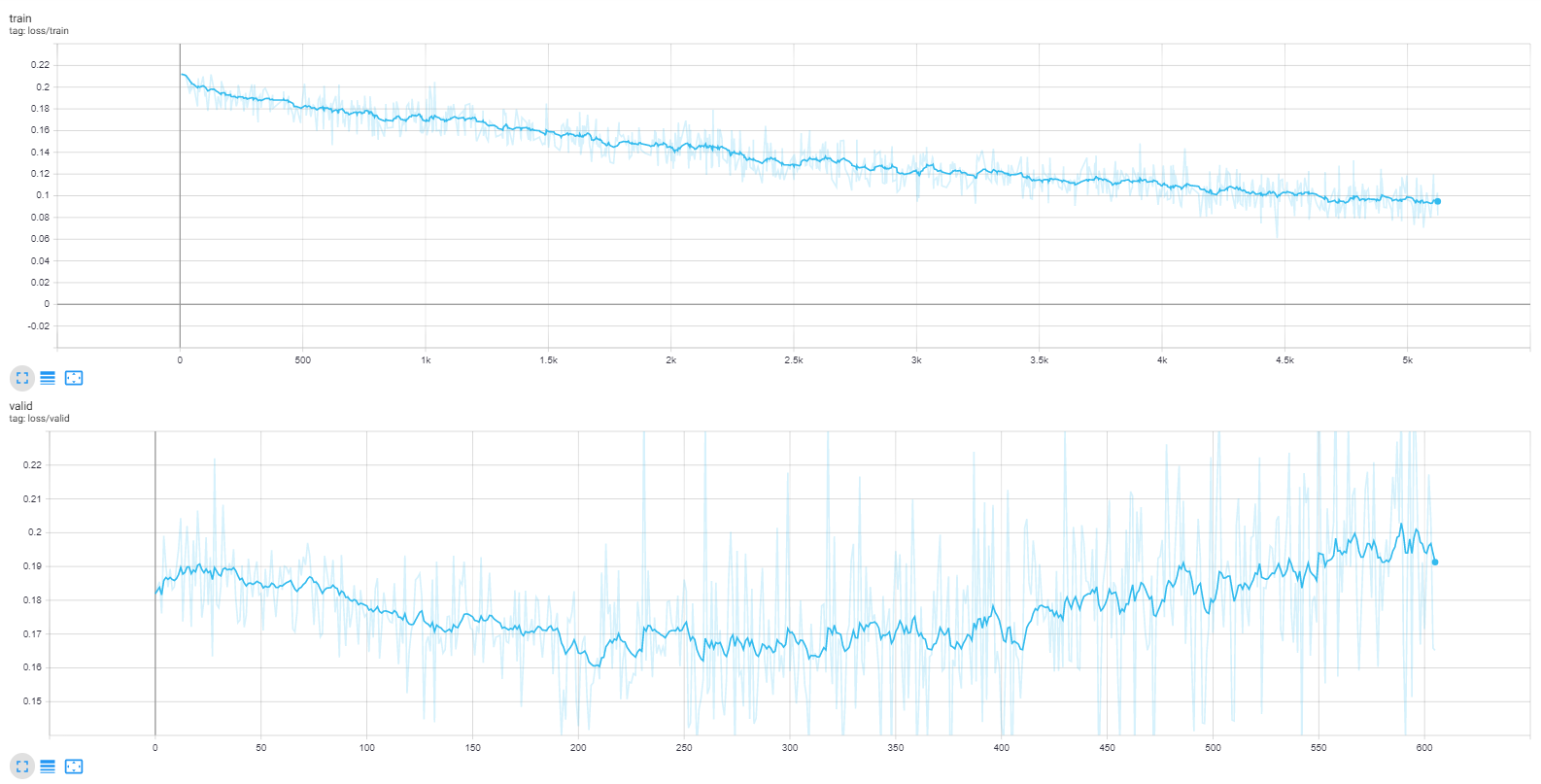}
    \caption{Overfitting example: One-hot encoding + LSTM.}
    \label{fig:overfitting_example}
\end{figure}

\end{document}